\documentclass[conference]{IEEEtran}
\IEEEoverridecommandlockouts
\usepackage{cite}
\usepackage{amsmath,amssymb,amsfonts}
\usepackage{algorithmic}
\usepackage{graphicx}
\usepackage{textcomp}
\usepackage{xcolor}
\usepackage{multirow}
\usepackage{url}

\newcommand{\jxc}[1]{{\color{black} #1}}

\def\BibTeX{{\rm B\kern-.05em{\sc i\kern-.025em b}\kern-.08em
    T\kern-.1667em\lower.7ex\hbox{E}\kern-.125emX}}
\begin{document}

\makeatletter
\newcommand{\linebreakand}{%
    \end{@IEEEauthorhalign}
    \hfill\mbox{}\par
    \mbox{}\hfill\begin{@IEEEauthorhalign}
}
\makeatother
    
\title{HRGCN: Heterogeneous Graph-level Anomaly Detection with Hierarchical Relation-augmented Graph Neural Networks}

% *Submission is double-blinded, comment out below author information

\author{\IEEEauthorblockN{Jiaxi Li}
\IEEEauthorblockA{\textit{University of Technology Sydney} \\
Sydney, Australia \\
jiaxi.li-1@student.uts.edu.au}
\and
\IEEEauthorblockN{Guansong Pang}
\IEEEauthorblockA{\textit{Singapore Management University} \\
Singapore 178902, Singapore \\
gspang@smu.edu.sg}
% \and
\linebreakand
\IEEEauthorblockN{Ling Chen}
\IEEEauthorblockA{\textit{University of Technology Sydney} \\
Sydney, Australia \\
ling.chen@uts.edu.au}
\and
\IEEEauthorblockN{Mohammad-Reza Namazi-Rad}
\IEEEauthorblockA{\textit{University of Wollongong} \\
Wollongong, Australia \\
mrad@uow.edu.au}
}
\maketitle

\begin{abstract}
This work considers the problem of heterogeneous graph-level anomaly detection. Heterogeneous graphs are commonly used to represent behaviours between different types of entities in complex industrial systems for capturing as much information about the system operations as possible. Detecting anomalous heterogeneous graphs from a large set of system behaviour graphs is crucial for many real-world applications like online web/mobile service and cloud access control. 
% Industrial systems are usually complex and mixed with many applications for different functionalities. Detecting anomalous system/application behaviours have become increasingly important in many industries. Through the communication among these applications, a significant number of API requests and logs are generated, enabling the detection of various anomalous events within the request traces and logs. Due to the complexities of objects \textcolor{red}{Ling: what do you mean by objects?}and their interactions in the traces, it is essential to consider as much information from the trace data to conduct efficient and effective anomaly detection. 
% Existing system/application trace anomaly detection approaches often bring additional log information alongside the trace structures to improve anomaly detection performance. 
To address the problem, we propose HRGCN, an unsupervised deep heterogeneous graph neural network, to model complex heterogeneous relations between different entities in the system for effectively identifying these anomalous behaviour graphs. HRGCN trains a hierarchical relation-augmented Heterogeneous Graph Neural Network (HetGNN), which learns better graph representations by modelling the interactions among all the system entities and considering both source-to-destination entity (node) types and their relation (edge) types.
% s in industrial systems.
% via pure trace structure data. 
% HRGCN trains a hierarchical relation-augmented Heterogeneous Graph Neural Network (HetGNN) to learn better graph representations by modelling the interactions among all the traces' source-to-destination neighbourhood types and edge types. 
% \textcolor{red}{Ling: HRGCN trains a hierarchical relation-augmented Heterogeneous Graph Neural Network (HetGNN), which learns better graph representations by modelling the interactions among all the traces and considering both source-to-destination node types and edge types. But before this, I think you need one line to explain first you model which data as heterogeneous graphs.}
% We further introduce a self-supervised learning module to perform Heterogeneous Graph Data Augmentation (HetGDA) to help train a more generalised HRGCN model.
% by producing a balanced AUC (Area under ROC) and AP (Average Precision). 
Extensive evaluation on two real-world application datasets shows that 
% 1) HRGCN achieves an AUC (area under ROC) of 0.86 , an AP (average precision) of 0.75 in microservices application data, and near-perfect AUC and AP in system flow data, respectively; \textcolor{red}{Ling: I don't think it is necessary to report the detailed performance of HRGCN here.} and 2) 
HRGCN outperforms state-of-the-art competing anomaly detection approaches.
% \textcolor{red}{Ling: it is confusing to mention pure-trace-based methods. What are pure-trace-based methods and non-pure-trace-based methods? Just say the state-of-the-art. Also, what about the effectiveness of HetGDA?} by 2\% - 20\%. In the end, 
We further present a real-world industrial case study to justify the effectiveness of HRGCN in detecting anomalous (e.g., congested) network devices in a mobile communication service. HRGCN is available at \url{https://github.com/jiaxililearn/HRGCN}.
% through vast trace events over time.
\end{abstract}


\section{Introduction}

% Anomaly detection is a critical problem in many industries, among which cyber-security is one of the most popular domains. Due to the complexity of industrial systems,  graph structures are often used to model the data to conduct various anomaly detection tasks. In particular, heterogeneous graphs are extensively explored in these applications due to their ability to capture rich information pertaining to relationships, multi-typed objects, and attributes. \textcolor{red}{Ling: there is no logical relation between this sentence and the following one...} Also, heterogeneous graphs'  ubiquity in many scenarios has recently attracted increasing research interest, such as academic graphs \cite{zhang2019heterogeneous}, user product review \cite{tang2021fraud}, recommendation \cite{liu2020heterogeneous}, as well as anomaly detection \cite{liu2019log2vec}. A popular graph anomaly detection methodology  is learning the graphs' embeddings to perform further detection tasks from the learnt representations \cite{dong2017metapath2vec, chang2015heterogeneous}. Therefore, the quality of the graph representations becomes a crucial foundation for anomaly detection tasks. In this context, heterogeneous graph neural networks (HetGNNs) are widely used for better quality when applied to heterogeneous graphs. It uses deep neural network layers to aggregate multi-typed information from all the neighbourhood objects to learn more distinct representations \cite{zhang2019heterogeneous}. 

\jxc{
Anomaly detection 
% plays a vital role in safeguarding systems from unexpected and potentially harmful events. It 
is an important area of data analytics and machine learning which involves the development of algorithms and techniques to identify deviations from normal behaviours.
% , enabling early detection and mitigation when anomalies appear.
By distinguishing between normal and abnormal patterns, anomaly detection empowers organisations to proactively address emerging issues, prevent system failures, and enhance reliability \cite{pang2021deep}. In the realm of large and complex systems, such as cloud platforms and mobile/IoT networks, anomaly detection has garnered increasing attention in recent years \cite{lindemann2021survey}. These systems are characterised by their extensive data volumes, diverse services and their relations, presenting formidable challenges in detecting anomalous patterns, such as system performance degradation, unauthorised access, or malicious activities. Addressing these challenges requires comprehensive evaluations of behaviours involving diverse entities in the system. In this context, behaviour data is typically represented by graphs to capture the entity interactions.
% complex software platforms.
% For example, 
Figure \ref{fig:example_network_data} illustrates an example of normal and abnormal devices based on heterogeneous mobile networks/graphs involving two types of entities (users and devices) and multiple types of interactions among the entities. Although the users may occasionally encounter intermittent ``bad" events, it does not necessarily imply that the entire graph exhibits abnormal behaviours. Therefore, a nuanced understanding of the complex inter-dependencies and interactions among the heterogeneous entities within the system is crucial for accurately detecting the anomalies.

% Specifically, anomaly detection in large and complex systems, such as cloud platforms and mobile/IoT networks, has gained significant popularity in recent years. Due to the vast amounts of data and numerous services, detecting anomaly patterns like overall performance degradation, unauthorised access, or malicious activities presents as challenging obstacles. In this work, we consider anomaly detection in complex software platforms via evaluating the full network behaviour. For example, Figure \ref{fig:example_network_data} illustrates one example of normal and abnormal devices based on network events. Although the users may experience intermittently with "bad" events, it may not result in the whole graph to be abnormal.
}

\begin{figure}
    \centering
    \scalebox{0.9}{
    \includegraphics[width=\linewidth]{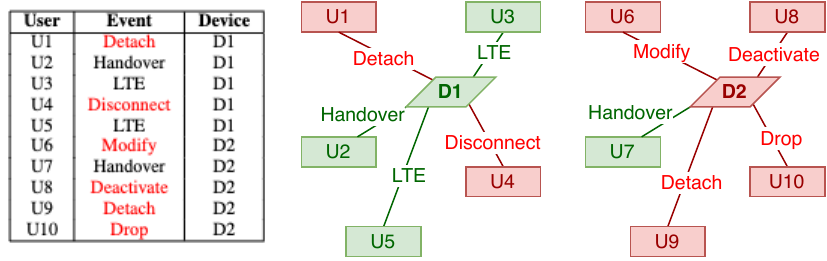}
    }
    \caption{Example mobile network events with heterogeneous entities and relations. Regular (green) vs. irregular (red) events and devices.}
    % in this setting.}
    \label{fig:example_network_data}
\end{figure}

There are three main streams of related anomaly detection approaches on graphs, including homogeneous-graph-based approaches \cite{ma2022deep,zhang2022dual,niu2023graph}, heterogeneous-graph-based approaches \cite{tang2021fraud,manzoor2016fast,zhang2019heterogeneous}, and hybrid approaches considering both graph and non-graph data (e.g., system logs) \cite{zhang2022deeptralog}. 
% While dealing with different types of graph data, t
Their common objective is to identify graph-level anomalies, i.e., exceptional graphs that deviate significantly from the other graphs. Compared with homogeneous-graph-based approaches that focus on dealing with graphs consisting of a single type of nodes and edges, heterogeneous-graph-based approaches generally perform better on complex-structured datasets. This is mainly because heterogeneous graphs, which can be used to represent behaviours between different types of entities in complex industrial systems, excel at capturing a variety of important information about the system operations for anomaly detection.
% by considering heterogeneous relationships. %but their representation capacity is limited as they ignore a number of important heterogeneous relation types.
% not provide the same performance results when running against more complex datasets close to real-world settings.
On the other hand, hybrid approaches like DeepTraLog \cite{zhang2022deeptralog} can learn normal patterns from more sources of data, such as heterogeneous graphs based on the system trace data and log embeddings based on the log texts.
% , for more effective anomaly detection.
% from the microservice application data.
% , which is a complex real-world application dataset. 
The hybrid approaches are generally more effective, but they rely on the availability of extra data sources, in addition to the heterogeneous graphs. Since these additional data sources are often unavailable or difficult to obtain, this work focuses on exploring the heterogeneous-graph-based approaches.
\jxc{Most current approaches \cite{zhang2019heterogeneous,wang2019heterogeneous,fu2020magnn,hu2020heterogeneous,zhao2020network,sun2021heterogeneous} 
% \gs{references needed}
based on heterogeneous graphs primarily focus on learning representations in a supervised or semi-supervised manner. However, in practical anomaly detection scenarios, acquiring high-quality labeled anomaly data is prohibitively costly.} There are two quick solutions to this problem. One is to apply an unsupervised anomaly detection optimisation objective, such as SVDD,  to replace the loss layer of the traditional HetGNNs \cite{zhang2019heterogeneous} so as to avoid the requirement for labelled anomaly data. Another solution is to adapt an unsupervised homogeneous-graph-based anomaly detection model, such as the recently proposed GLocalKD \cite{ma2022deep}, by replacing the used homogeneous GNNs with HetGNNs layers. Nonetheless, both of the approaches could not produce competitive results in the context of anomaly detection within complex systems (see Sec. \ref{subsec:results}). 
% This work is also motivated by solving system/application anomaly detection tasks in real-world industrial settings. One of the challenges is the lack of labelled data when dealing with real-world industrial datasets. 
% Therefore, we focus on an unsupervised learning methodology to address these situations and settings of industries. 

% In this scenario, we found that the traditional Heterogeneough Graph Neural Network does not perform well under unsupervised circumstances. Thus, in this work, we'll
To this end, we propose Hierarchical Relation-augmented Graph Convolution Neural Network (HRGCN), which is an unsupervised approach that extends heterogeneous HetGNNs by exploiting more complex heterogeneous relation hierarchies for identifying abnormal graphs that are difficult to spot by considering only shallow heterogeneous relations. 
Figure \ref{fig:HRGCN_diff} provides a simple example to illustrate how anomalous and normal behaviours can be distinguished with fine aggregation based on node types. Particularly, we first introduce a hierarchical relation module that considers deeper interactions within the source-destination neighbourhood types and edge types rather than considering all neighbour types and edge types simultaneously. Then, we introduce a self-supervised HetGNN heterogeneous relation prediction module that does not require human-annotated anomaly information.  Furthermore, to support the self-supervised task, we introduce a Heterogeneous Graph Data Augmentation (HetGDA) method to enhance the data coverage during the training process and produce a more generalised model. Finally, the model is trained by using a joint objective of one-class classification and self-supervised heterogeneous relation prediction. 
% Overall, we combine all the proposed modules onto the heterogeneous graph neural network and aggregate the intermediate embeddings and outputs from these modules to produce the final graph representation and obtain an anomaly score from the combination of SVDD distance and self-supervised prediction.

\begin{figure}
    \centering
    \includegraphics[width=\linewidth]{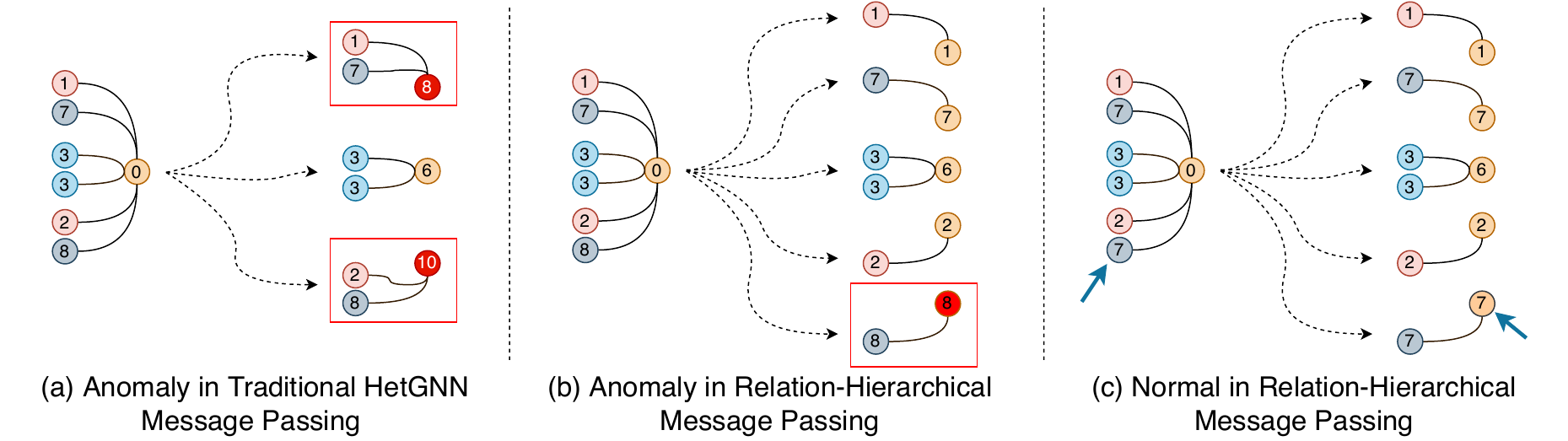}
    \caption{Simple Example Illustrates the Architectural Difference in Message Passing (MP). Each node is labelled with node attribute value which is passed onto its neighbour during MP. Traditional HetGNN (a) considers all the neighbour nodes with mixed node types at once without further specification. In contrast, the proposed model (b, c) considers each source-destination pair individually during MP. Assume simple criteria where a value greater than 7 is anomalous. (a) and (b) identify the graph as an anomaly since anomalous behaviours are detected. (c) is a normal graph introduced with a different node in the neighbourhood, the traditional HetGNN would still identify it as an anomaly, while the proposed architecture can effectively avoid such errors.}
    % in this setting.}
    \label{fig:HRGCN_diff}
\end{figure}

In summary, this work makes the following contributions:
\begin{itemize}
    \item We propose a novel hierarchical heterogeneous graph convolutional network model, i.e., HRGCN, for improved representation learning on heterogeneous graph-level anomaly detection. HRGCN can capture the general structural and attributed heterogeneity and is extended to the closer relations within source-to-destination neighbourhood types and edge types, which makes it effective in learning complex system-flow-like graphs for anomaly detection.

    \item We introduce a self-supervised prediction module and heterogeneous graph data augmentation method for training HRGCN more effectively in unsupervised settings.
    % improve the training process further and provide a more generalise performance to the trained model.

    \item We conduct extensive experiments on two system-focused graph datasets with distinct structural properties and attributes. The result demonstrates significant performance improvement in HRGCN compared to other state-of-the-art baseline models.

    \item We conduct extensive case studies over real-world-like mobile network event data. HRGCN demonstrates a superior performance running in a real-world industrial setting compared to other state-of-art baseline models.
\end{itemize}

% \jx{A1: The paper is structured as follow. Section \ref{related_work} introduces the related work around Heterogeneous GNN and graph-level anomaly detection. Section \ref{method} illustrates the detailed methodologies and architecture of the proposed HRGCN. Section \ref{experiments} provides detailed experiments, implementation, and performance discussion over the results. Section \ref{case_study} conducts an industrial case study by applying HRGCN on a real-world dataset and scenario. Finally, our conclusion is presented in Section \ref{conclusion}.}

\section{Related Work}\label{related_work}

\textbf{Heterogeneous Graph Neural Networks.} Heterogeneous Graphs (HGs) have been widely studied in representation learning \cite{wang2022survey}. Dong et al. introduced a heterogeneous graph model, metapath2vec, to produce scalable network representations \cite{dong2017metapath2vec}. Shi et al. created the HERec model specifically for recommendation systems with heterogeneous graph data and introduced an extended matrix factorisation (FM) module to optimise the rating prediction tasks \cite{shi2018heterogeneous}. Hu et al. introduced a three-way characterisation to combine object embeddings with meta-path-based context embedding for the recommendation tasks \cite{hu2018leveraging}, and further introduced an adversarial learning model, HeGAN, on heterogeneous graphs in 2019 \cite{hu2019adversarial} which is inspired by the generative adversarial networks(GANs).

Graph Neural Networks (GNNs) \cite{scarselli2008graph} are neural network models that can be used directly to analyse graph structure data and focuses on revealing complicated relationships between objects in a network. In the past few years, research has been conducted on improving GNNs and applying them to different areas. A standard learning methodology from GNNs generates meaningful representations describing the input graphs. To this extent, graph-structured data demonstrates the capability to easily model more complicated systems, such as social networks \cite{fan2019graph}, biomedical networks \cite{yue2020graph},  the Internet \cite{zhang2021graph}, and other complex information networks with attributes \cite{zhou2020graph}. Therefore, producing helpful network representations becomes a continuous research topic. Li et al. proposed the Gated Graph Neural Network (GGNN), which uses a gated recurrent unit (GRU) to remember or forget messages during each message passing in GNNs \cite{li2015gated}. Zhang et al. proposed the Heterogeneous Graph Neural Network (HetGNN), which incorporates heterogeneous structural information, attributes, and contents \cite{zhang2019heterogeneous}. The HetGNN model is aimed to produce helpful node embeddings within the networks by introducing multiple aggregation modules based on various heterogeneous information. Wang et al. proposed the Heterogeneous Graph Attention Network (HAN) and introduced the hierarchical attention mechanism into the GNNs \cite{wang2019heterogeneous}. Specifically, it includes both node-level and semantic-level attention, which considers the importance between the nodes and meta-paths and the importance within different meta-paths. Fu et al. proposed a meta-path aggregated Heterogeneous GNN model (MAGNN) to overcome the limitations of traditional meta-path-based GNNs \cite{fu2020magnn}. It introduces the additional node content transformation, intra-meta-path aggregation, and inter-meta-path aggregation to accommodate input node attributes and the message combination between meta-paths. \jxc{Hu et al. proposed the Heterogeneous Graph Transformer to model heterogeneity over each edge and feed the sampled graph features into a Transformer before the message passing \cite{hu2020heterogeneous}}. Zhao et al. proposed a Heterogeneous Network Embedding based on network schema (NSHE) \cite{zhao2020network}. Instead of using the meta-paths, utilising the network schema directly could preserve rich semantics from the high-order structure when learning the network representations. \jxc{Similarly, Sun et al. introduced a graph classification method using hypergraph embeddings, which is a network-based learning framework for  heterogeneous hypergraphs \cite{sun2021heterogeneous}}.
% More HetGNN examples? Such as Self-supervised GNN, HeCo, etc.

\textbf{Graph-level Anomaly Detection.} There are limited research on anomaly detection via Heterogeneous Graph representation learning, particularly detecting anomalous behaviours from running systems and applications. Manzoor et al. proposed a streaming low-latency model to detect anomalous actions from the heterogeneous graphs constructed from system APIs \cite{manzoor2016fast}. It introduces a light-weighted hashing mechanism in real-time to update and compute the similarities of the hashed graphs. MultimodalTrace \cite{nedelkoski2019anomaly} is a graph-based anomaly detection model and leverages a multi-modal LSTM module to learn the graph representations from only the normal graphs. TraceAnomaly \cite{liu2020unsupervised} is an unsupervised model proposed to detect anomaly traces from microservice applications. It considers only service-level graphs and uses the variational auto-encoders (VAE) to detect anomaly patterns. Zhang et al. \cite{zhang2022deeptralog} proposed a DeepTraLog model to combine both log data and graph-structured data to detect anomalies from microservice applications more confidently. Another closely related line of research is homogeneous graph anomaly detection \cite{ma2022deep,zhang2022dual,niu2023graph}. They also perform graph-level anomaly detection, but they are focused on homogeneous graphs. Thus, they cannot work effectively when applied to heterogeneous graphs. We also demonstrate that simple adaptation of these homogeneous graph approaches cannot work well in our experiments.
% Due to the difficulty of capturing abnormalities at multiple scales of a graph, 

\section{Methodology}\label{method}

% \jx{Done: Adding a table summarise the notions used in Methodology}
\jxc{We first introduce a baseline model that adapts a popular deep one-class classifier, Deep SVDD \cite{ruff2018deep}, for heterogeneous graph-level anomaly detection. We then introduce the proposed framework and methodologies which are related to the proposed HRGCN model in details. A list of notations used is summarised in Table \ref{tab:notion_sum}.}
% summarises references of the frequently used notions in this paper.}

\begin{table}
  \caption{Notion Summary}
  \label{tab:notion_sum}
  \scalebox{0.9}{
  \begin{tabular}{l|l}
    % \toprule
    \hline
    \textbf{Notions} & \textbf{Definitions} \\
    \hline
    % \midrule
    \texttt{$V,E,T_v,T_e$}& Collection of node, edge, node type, edge type \\
    \texttt{$v_i, e_{j,i}$}& Node $i$, edge between node $j$ and $i$ \\
    \texttt{$T_{v_j}$} & Node type of node $j$\\
    \texttt{$t_{src}, t_{dst}$} & Source node type, destination node type\\
    \texttt{$t_{e_{j,i}}; t \in T_e$} & Edge type of the edge between node $j$ and $i$\\
    \texttt{$\mathbf{c}$} & SVDD center\\
    \texttt{$Batch(i)$} & A collection of graphs in the $i$-th batch\\
    \texttt{$n_{B}$} & The batch size\\
    \texttt{$\mathcal{N}(i)$} &  Neighbours of node $i$\\
    \texttt{$\mathcal{G}_i$} & Embedding of graph $i$\\
    \texttt{$s(\mathcal{G}_i)$} & Anomaly score of graph $i$\\
    \texttt{$\tilde{\mathbf{T}_v}$} & A collection of augmented node types of graph\\
    \texttt{$\tilde{\mathbf{T}_e}$} & A collection of augmented edge types of graph \\
    \texttt{$\tilde{\mathbf{E}}$} & A collection of augmented edges\\
   
    % \texttt{HRGCN}& 0.855 &	\textbf{1.0} & \textbf{0.762}	 &	\textbf{1.0}\\
    \hline
    % \bottomrule
  \end{tabular}
  }
\end{table}

\subsection{Baseline Model: One-Class Heterogeneous Graph Convolution Network} \label{section:baseline_hetgcn}
The original HetGNN model \cite{zhang2019heterogeneous} was built to learn on a large graph to generate each node's representations. It was implemented based on the neighbour sampling mechanisms and LSTM layers for encoding node/neighbourhood contents. Furthermore, the loss function is computed by calculating the cross entropy loss between the triplet outputs, including positive and negative representations. To comply with our settings, we have extended the HetGNN model with Message Passing (MP) \cite{gilmer2017neural} and One-Class SVDD loss \cite{ruff2018deep}.

\subsubsection{Heterogeneous Convolution Neural Network (GCN)}
An attributed heterogeneous graph can be defined as $G = (V, E, T_V, T_E)$, where $v_i \in V$ denotes the nodes, $e_{j,i} \in E$  denotes the edges, and $T_V$ and $T_E$ represent the sets of node types and edge types respectively. For each node $v_i$, it associates with a set of node attributes $\mathbf{x}_i \in \mathbb{R}$ for node $i$, and its neighbourhood can be defined as $\mathcal{N}(i)$. On the other hand, message passing can also be expressed as neighbourhood aggregation. Under this implementation, message passing at $k$th message propagation in graph neural network can be described as Eq. \ref{eq:message_passing}:

\begin{equation} \label{eq:message_passing}
    \resizebox{0.9\hsize}{!}{$\mathbf{x}_i^{(k)}=\gamma^{(k)}\left(\mathbf{x}_i^{(k-1)}, \sum_{j \in \mathcal{N}(i)} \phi^{(k)}\left(\mathbf{x}_i^{(k-1)}, \mathbf{x}_j^{(k-1)}, \mathbf{e}_{j, i}\right)\right).$
    }
\end{equation}
where $\phi$ and $\gamma$ are differentiable functions \jxc{for constructing messages and updating node embeddings}.
Furthermore, the implemented Graph Convolution Network (GCN) layer can be mathematically described as below 

\begin{equation}\label{eq:GCN}
    \resizebox{0.9\hsize}{!}{
    $\mathbf{x}_i^{(k)}=\sum_{j \in \mathcal{N}(i) \cup\{i\}} \frac{1}{\sqrt{\operatorname{deg}(i)} \cdot \sqrt{\operatorname{deg}(j)}} \cdot\left(\mathbf{w}^{T} \cdot \mathbf{x}_j^{(k-1)}\right)+\mathbf{b}.$
    }
\end{equation}
That is, the node features of all the neighbour nodes are first transformed by a weight matrix $\mathbf{w}$, then normalised by the degrees of the node pairs, and the final node feature is acquired by summing up the transformed values of all the neighbours at the end.

When considering the heterogeneous node types, each node type $t$ from the neighbourhood is transformed using a different neighbourhood encoder $\phi_t$ to transform the contents from all the nodes in the same neighbourhood type. Eq. \ref{eq:HetGCN_node_type} and Eq. \ref{eq:HetGCN} below describe the masked message passing processes with different node types:

\begin{equation}\label{eq:HetGCN_node_type}
    \resizebox{0.9\hsize}{!}{
    $\mathcal{F}(x_i, x_j, t)= \begin{cases} 
    0, \text{ if } t_{v_j} \neq t \\
    \frac{1}{\sqrt{\operatorname{deg}(i)} \cdot \sqrt{\operatorname{deg}(j)}} \cdot\left(\mathbf{w}^{T} \cdot \mathbf{x}_j^{(k-1)}\right)+\mathbf{b}, \text{ else }
    \end{cases}
    $
    }
\end{equation}

\begin{equation}\label{eq:HetGCN}
    \resizebox{0.9\hsize}{!}{
    $\mathbf{x}_i^{(k)}={Concat}_{t \in T_{v_j}}\left(\sum_{j \in \mathcal{N}(i) \cup\{i\}} \phi_t\left(\mathcal{F}(x_i, x_j, t)\right)\right).$
    }
\end{equation}

The neighbour nodes of the different types are fed into different differentiable functions. Then concatenation is performed on all the transformed neighbourhood types to get resulting node representations.
 
In the end, to produce the embedding of the graph $\mathcal{G}$, max pooling is performed on all the node representations from the last message passing iteration $K$ to produce a single vector for the final representation: $\mathcal{G} = \max_{i \in V}\left(x_i^{(K)}\right)$.
% \begin{equation}\label{eq:HetGCN_graph}
%     \mathcal{G} = \max_{i \in V}\left(x_i^{(K)}\right)
% \end{equation}

\subsubsection{One-Class Deep SVDD} \label{section:svdd}
Deep Support Vector Data Description (Deep SVDD) is an unsupervised loss function used in the area of Anomaly Detection. SVDD can be described as a quadratic loss between the network representation $\phi(\mathcal{G}_i;\mathcal{W})$ and a hyper-sphere center $\mathbf{c}$. 
% I.e., Eq. \ref{eq:svdd} denotes the objective function, and the goal
The specific objective is to minimise the distance of every network representation to the hyper-sphere center:

\begin{equation}\label{eq:svdd}
    \min _{\mathcal{W}} \frac{1}{n} \sum_{i=1}^n\left\|\phi\left(\mathcal{G}_i ; \mathcal{W}\right)-\boldsymbol{c}\right\|^2+\frac{\lambda}{2} \sum_{\ell=1}^L\left\|\boldsymbol{W}^{\ell}\right\|_F^2.
\end{equation}
where $\lambda$ is a regularisation parameter and $\mathbf{W}^l$ denotes the weight of the layer $l$ in the HetGNN model.
After training the one-class HetGNN, given a test graph $\boldsymbol{g} \in \mathcal{G}$, the anomaly score $s$ is then defined as the distance to the fixed SVDD center $\mathbf{c}$, i.e., $s(\mathcal{G}_i) = \left\|\phi\left(\mathcal{G}_i ; \mathcal{W}\right)-\boldsymbol{c}\right\|^2$. \jxc{Note that, the center $\mathbf{c}$ is computed as an averaged vector of the graph embeddings within the first data batch:}

\begin{equation}
    \mathbf{c} = \frac{\sum_{i=1}^{n_{B}}\left(\phi(\mathcal{G}_i)\right)}{n_B}, \{\mathcal{G}_i \in Batch(1)\}.
\end{equation}

% \jx{Done: Add details on how the fixed center is calculated.}

\subsection{Modelling Hierarchical Heterogeneous Relations in Graphs of Normal Class}

The above-implemented HetGCN model aggregates all the neighbourhood types together regardless of the source node type. In scenarios where there are numerous source node types, the exchange of information between specific source-destination node types may diminish in subsequent transformations and aggregations within the network architecture. To address this issue and preserve more information during the learning process, we propose the HRGCN, which introduces a segregation of the architecture based on individual source nodes and edge types. This approach allows for the incorporation of deeper hierarchical relations within the heterogeneous graph. Figure \ref{fig:architecture} illustrates the overall framework of HRGCN in a detailed example, including the relation-hierarchy feature convolution module and self-supervised module.

\begin{figure*}
      \centering
      \scalebox{0.9}{
        \includegraphics[width=\textwidth]{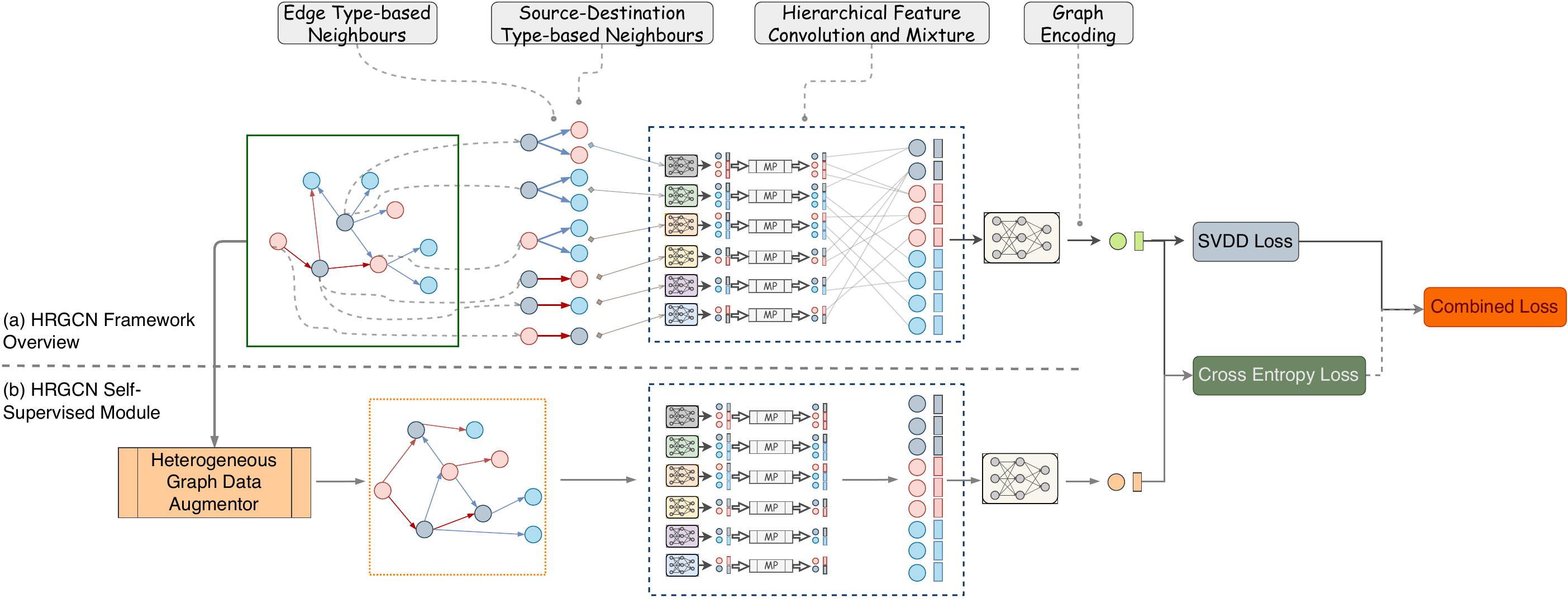}
      }
      \caption{(a) Overall Architecture of HRGCN. It first identifies all the edge types and source-to-destination node types in the input graph and encodes the segregated node features via hidden NN layers. Then the encoded node features in each edge and source-to-destination type are fed to the message-passing network to aggregate the neighbour information. Next, the aggregated neighbour features are then mixed and aggregated for each node in the graph. The final node embeddings are aggregated to create graph representation via the final NN layers and the SVDD loss. (b) The Self-Supervised module framework. It augments the input HGs and produces the graph representation with the HRGCN architecture, and then both the input graph and augmented graph are combined via Cross-Entropy (CE) Loss. In the end, the final loss of the input graph data is calculated based on the SVDD error and weighted CE errors.}
      \label{fig:architecture}
\end{figure*}

\subsubsection{Source-Destination Node Type and Edge Type Relations}

To establish a more distinctive hierarchy during the message passing process from various source nodes, we introduce the inclusion of source node type $t_{src}$ and destination node type $t_{dst}$, where $t_{src} \in T_v$ and $t_{dst} \in T_v$, when iterating within the network. 

The source-destination hierarchical masking function $\mathcal{F}^{1}$ can then be extended as below

\begin{equation}\label{eq:HRGCN_src_mask}
\scalebox{0.9}{
    $\mathcal{F}^{1}(i, j, T_v)= \sum_{t_{src} \in T_v, t_{dst} \in T_v}\left(\begin{cases} 
    0, \text{ if } (t_{v_i},t_{v_j}) != (t_{src},t_{dst})\\
    Norm(x_i, x_j), \text{ else }
    \end{cases}\right),$
    }
\end{equation}

where $Norm(x_i, x_j)$ is a normalisation step of the node pairs for each source-destination type which is defined as follows
\begin{equation}\label{eq:degnorm}
    \resizebox{0.9\hsize}{!}{
    $Norm(x_i, x_j) = \frac{1}{\sqrt{\operatorname{deg}(i)} \cdot \sqrt{\operatorname{deg}(j)}} \cdot\left(\mathbf{w}^{T} \cdot \mathbf{x}_j^{(k-1)}\right)+\mathbf{b}.$
    }
\end{equation}

% Eq. \ref{eq:degnorm} describes the normalisation step of the node pairs for each source-destination type.

With an additional hierarchy introduced for source node types, the differentiable functions are expanded to accommodate the resulting changes. The message passing graph can then be defined as

\begin{equation}\label{eq:HetGCN_src}
    \resizebox{0.9\hsize}{!}{
    $\mathbf{x}_i^{(k)}={Concat}_{t \in T_{v}}\left(\sum_{j \in \mathcal{N}(i) \cup\{i\}} \phi_{t_{i,j}}\left(\mathcal{F}^{1}(i, j, T_v)\right)\right).$
    }
\end{equation}
where $\phi_{t_{i,j}}$ denotes as individual differentiable function of each source-destination node type pair.

A similar hierarchical enhancement can also be applied to each edge type relation, further segregating the hierarchy by combining edge types and source-destination node type pairs via Eq. \ref{eq:HetGCN_edge_mask}.

\begin{equation}\label{eq:HetGCN_edge}
    \resizebox{0.9\hsize}{!}{
    $\mathbf{x}_i^{(k)}={Concat}_{t \in T_{v};T_{e}}\left(\sum_{j \in \mathcal{N}(i) \cup\{i\}} \phi_{t_{i,j}}\left(\mathcal{F}^{2}(i, j, T_e)\right)\right),$
    }
\end{equation}
\begin{equation}\label{eq:HetGCN_edge_mask}
    \resizebox{0.9\hsize}{!}{
    $\mathcal{F}^{2}(i, j, T_e)= \sum_{t \in T_e}\left(\begin{cases} 
    \phi_{t_e}\left(\mathcal{F}^{1}(i, j, T_v)\right), & \text{if }t_{e_{i,j}} = t \\ 
    0 & \text{else}
    \end{cases}\right).$
    }
\end{equation}

where $\phi_{t_e}$ is the differentiable function for each edge type in the graph. Furthermore, the final node representations are updated as Eq. \ref{eq:HetGCN_edge} which aggregates on all the source-destination node type pairs $T_{v_{i,j}}$ and edge types $T_e$. This indicates that for edges in each edge type, we will perform the message passing for all the source-destination node type pairs. 

% \jxc{Note that, when separating the source-destination node type relations and edge relations, message passing (MP) is performed at each granular relation type. Although the sparsity of the segments is increased, the significance of each separated relation type or type-pair are combined to retain a more distinctive representation of the corresponding nodes and graph.}

% \jx{Done: address the sparsity of the architecture and its impact}

% \subsubsection{Vectorisation}
% To optimise the computation efficiency. Write it later if needed.

\subsection{Heterogeneous Graph Data Augmentation for Self-Supervised Normality Learning}
Graph Data Augmentation (GraphDA) methods are employed to augment graph datasets and enhance the representation learning process. Existing graph augmentation approaches \cite{ding2022data} primarily focus on homogeneous graphs and simple heterogeneous graphs. However, to cater to attributed heterogeneous graphs, we introduce four heterogeneous graph augmentation methods specifically designed to complement the proposed HRGCN self-supervised training process.

\subsubsection{Heterogeneous Edge Perturbation}
Edge Perturbation is widely used in GraphDA methods in different Deep Graph Learning tasks \cite{
%velickovic2019deep,
you2021graph,you2020graph,zhu2021graph}. It is the technique of randomly adding or removing edges within the graph. Given an adjacency matrix, edge perturbation preserves the original node order and overwrites a portion of the edges. For example, Eq. \ref{eq:edge_perturbation} describes the process of perturbation of edges. By obtaining a randomly sampled corruption matrix $\mathbf{C}$, then XOR operation $\oplus$ is carried out between the adjacency matrix $\mathbf{A}$ and the corrupted matrix $\mathbf{C}$ to create a new adjacency matrix $\tilde{\mathbf{A}}$:

\begin{equation}\label{eq:edge_perturbation}
    \scalebox{0.9}{$\tilde{\mathbf{A}}=\mathbf{A} \oplus \mathbf{C}.$}
\end{equation}

To also preserve the heterogeneity from the original graphs and generate better quality heterogeneous graphs, the matrix $\mathbf{C}$ is generated based on the prior information of the source-destination node type pair and edge type distributions:

\begin{equation}
    \scalebox{0.9}{$\mathbf{C}(T_{v_i}, T_{v_j}, T_e) \sim \left\{P(T_{v_i}, T_{v_j}), P(T_e)\right\}.$}
\end{equation}

\subsubsection{Heterogeneous Edge Replacement}
Edge Replacement is an alternative edge augmentation method involves addition and removal of edges from the original graph. In contrast to the randomness of Edge Perturbation, which applies to all possible node pairs, edge replacement maintains the total number of edges in the specific neighbourhood while introducing new edges to the graph. Moreover, the selection of edges to add follows a criterion based on their rarity within the graphs. Edges with infrequent occurrences in the neighborhood are more likely to be generated in the augmented graphs. The replacement can be described as follows:

\begin{equation} \label{eq:edge_add}
    \scalebox{0.9}{$\mathbf{R} \sim P(e_{i,j}) \propto \frac{1}{P(T_{v_i}, T_{v_j}, T_e)}.$}
\end{equation}

In addition, some existing edges in the same neighbourhood are removed to form similar-sized resulting graphs, as described in Eq. \ref{eq:edge_replace}. The new set of edges $\tilde{\mathbf{E}}$ is acquired by replacing a subset of edges $\mathbf{E`}$ in the neighbourhood with the newly sampled edges $\mathbf{R}$.
\begin{equation} \label{eq:edge_replace}
    \scalebox{0.9}{$\tilde{\mathbf{E}} = \mathbf{E} \cup \mathbf{R} \setminus \{\mathbf{E`} \subseteq \mathbf{E}(T_{v_i}, T_{v_j}, T_e)\}.$}
\end{equation}

\subsubsection{Heterogeneous Node and Edge Type Swapping}

Another approach to enhance the training process of HRGCN is by creating augmented graphs through node and edge type swapping. In the node type swap augmentation step, a random node type pair $(\mathbf{T}_{v_{a}}, \mathbf{T}_{v_{b}})$ is selected for the swap operation, but only if $\mathbf{T}_{v_{a}} \neq \mathbf{T}_{v_{b}}$. Consequently, a new set of node types, denoted as $\tilde{\mathbf{T}_v}$, can be defined after the swapping process.

\begin{equation}\label{eq:node_swap}
    \scalebox{0.9}{$\tilde{\mathbf{T}_v} \sim \{\mathbf{V_a^`} \subseteq \mathbf{V}(\mathbf{T}_{v_{a}}), \mathbf{V_b^`} \subseteq \mathbf{V}(\mathbf{T}_{v_{b}})\}.$}
\end{equation}

On the other hand, edge type swap can be performed using a similar method as shown in Eq. \ref{eq:edge_swap} below. This operation results in a new edge type list denoted as $\tilde{\mathbf{T}e}$, which is generated based on the edge type pair $(\mathbf{T}{e_{a}}, \mathbf{T}{e{b}})$.

\begin{equation}\label{eq:edge_swap}
    \scalebox{0.9}{$\tilde{\mathbf{T}_e} \sim \{\mathbf{E_a^`} \subseteq \mathbf{E}(\mathbf{T}_{e_{a}}), \mathbf{E_b^`} \subseteq \mathbf{E}(\mathbf{T}_{e_{b}})\}.$}
\end{equation}

% \jxc{In the end, the newly created collections of node types $\tilde{\mathbf{T}_v}$ and edge types $\tilde{\mathbf{T}_e}$ will replace the original graph attributes and serve as the augmented data in the self-supervised module.}

% \jx{Done: Revise swapping technique to make it more clear.}

\subsubsection{Self-Supervised Loss with HetGDA}
The augmented graph data is subsequently employed in the self-supervised module as depicted in Figure \ref{fig:architecture} (b). Within the self-supervised context, the augmented graph data is treated as abnormal graphs when computing the supervised loss and passed through the same HRGCN architecture during the training process. 
% In our methodology, Cross-Entropy Loss and Deviation Loss \cite{pang2019deep} are being used on different datasets in this study.

\subsection{Training and Inference}

By incorporating both SVDD and self-supervised modules into the model, the overall objective is defined as minimizing the following function:

\begin{equation}\label{eq:final_objective}
    \left(\frac{1}{n} \sum_{i=1}^n\left\|\phi\left(\mathcal{G}\right)-\boldsymbol{c}\right\|^2+\frac{\lambda}{2} \sum_{\ell=1}^L\left\|\boldsymbol{W}^{\ell}\right\|_F^2 + \alpha\sum_{i=1}^{n}\mathcal{S}(\mathcal{G}, \tilde{\mathcal{G}})\right).
\end{equation}
where $\mathcal{S}(\mathcal{G}, \tilde{\mathcal{G})}$ represents the self-supervised objective using the augmented graphs and labels, and $\alpha$ represents the weight assigned to the self-supervised loss in the module. I.e., $\mathcal{S}(\mathcal{G}, \tilde{\mathcal{G})}=$
\jxc{
\begin{equation}\label{eq:self_supervise_obj}
\begin{split}
    -\frac{1}{N}\sum^{N}_{g\in\{\mathcal{G} \cup \tilde{\mathcal{G}}\}}(y\log(\phi(g)) + (1-y)\log(1-\phi(g))).
\end{split}
\end{equation}
}
The training iteration comprises 3 steps, 1) compute SVDD loss from the graph representation and centre $\mathbf{c}$ \jxc{as described in Section \ref{section:svdd} }; 2) compute self-supervised loss between input graph $\mathcal{G}$ and augmented graph $\tilde{\mathcal{G}}$ \jxc{as Eq. \ref{eq:self_supervise_obj}}; 3) sum SVDD and self-supervised losses adjusted by a weight parameter $\alpha$ \jxc{as equation \ref{eq:final_objective}}.
In the end, the anomaly score is obtained as below where output graph representation is denoted as $\phi_1$ while the self-supervised prediction score is represented by $\phi_2$.

\begin{equation}\label{eq:final_score}
    s(\mathcal{G}_i) = \left\|\phi_1\left(\mathcal{G}_i\right)-\boldsymbol{c}\right\|^2 *  \phi_2(\mathcal{G}_i)
\end{equation}

\section{Experiments}\label{experiments}

\subsection{Datasets}\label{sec:dataset}
% To ensure the experiment results reflect the model performance closer to the real-world scenarios, only system and application API datasets are used to get the optimal comparison results of the feasibility of the proposed model.

% There are t
Despite its broad application domains, there are limited publicly accessible datasets for heterogeneous graph-level learning tasks. Our experiments use two real-world datasets with heterogeneous system and application API entities.
Even though labels are provided with both datasets, an unsupervised learning setting where training data contains only normal graphs is conducted for all the experiments to mimic real-life application scenarios where abnormal data is difficult and/or costly to obtain.
% unlabelled in industrial settings. 
% Furthermore, the training datasets are created using only the normal graphs, while abnormal graphs are used in evaluation and testing.

\noindent\textbf{Train Ticket Graph Dataset (TraceLog).}
TraceLog is a large-scale heterogeneous graph dataset derived from a train ticket-based microservice system \cite{zhou2018fault
%,zhou2018poster
}
% is a medium-scale open-source micro-service application
for booking training tickets.
% and has been widely used in research of micro-service architectures and their
It is widely used for evaluating anomaly detection in system trace log sequences \cite{zhou2019latent, zhao2021identifying, liu2020unsupervised}. TraceLog has four fault classes and 14 fault subclasses for different faulty types, such as asynchronous interaction, multi-instance, configuration-related, and monolithic \cite{zhang2022deeptralog}. All 14 system faults are used as anomalies against normal system traces.
% \jxc{The detailed training and testing data setup is also described in Supplement \ref{sup:tracelog_details}.}

% Details are also described in Table \ref{tab:dataset_overview}. 
% \gs{information about the types of nodes and edges is needed, e.g., the total number of node types and edge types, the average number of node and edge types per graph, average number of graph nodes per graph, etc.} \jx{Details added.}
% . We also sample 65,000 normal graphs from the full dataset 
% while the teest data
% are used during the training process. The dataset is split into three subsets, training, evaluation, and testing sets. 65,000 normal graphs for the training set and ~33,667 mixed graphs(
% contains 22,000 normal graphs and 11,667 anomalous graphs. We further extract 65,000 normal graphs from the
% ) for evaluation and testing sets, respectively. 

% Each graph in the dataset contains around 900 nodes and \textcolor{blue}{X} edges. 

\noindent\textbf{System Flow Graph Dataset (FlowGraph).} FlowGraph is a system flow graph dataset created and used in\cite{manzoor2016fast}, where each graph consists of heterogeneous user web browsing activities, such as watching YouTube videos, downloading files and artifacts, opening \textit{cnn.com} webpage, opening Gmail, and playing video games.
% \gs{what are nodes, and edges?}\jx{Added below.}
The dataset contains 600 graphs, including one attack and five normal scenarios, with 100 graphs in each scenario. The attack scenario is categorised as drive-by download actions triggered by visiting unsecured URLs that exploit malicious software to gain root access.

% Specifically, FlowGraph has fewer graphs and types of nodes than TraceLog, but each graph is much larger and contains around 8,411 nodes and 12,730 edges.
\jxc{
% Supplement \ref{sup:dataset_overview}provides
A detailed introduction of the two datasets, including graph statistics and train/test configurations, can be found in Appendix \ref{sup:dataset_overview}.}
% \gs{add a table here to describe key statistics of these two datasets, something like table 4.}\jx{Added in Table \ref{tab:dataset_overview}}

\subsection{Competing Methods and Evaluation Metrics}
% The initial experiments are conducted to compare 
Our proposed model HRGCN is compared with four state-of-the-art graph-level anomaly detection models from diverse related research lines, including StreamSpot \cite{manzoor2016fast}, DeepTraLog \cite{zhang2022deeptralog}, GLocalKD \cite{ma2022deep}, and one-class HetGCN (OCHetGCN) \cite{zhang2019heterogeneous,ruff2018deep}.
% on heterogeneous graph data. Due to the data's heterogeneity, some models are updated with adaption to the heterogeneous graph data. In addition, we are restricting the training set to only including normal graphs. Therefore, the comparison models are modified accordingly to accommodate these settings.

\textbf{StreamSpot} is a seminal work on heterogeneous graph anomaly detection.
% Unlike the other three comparison models, StreamSpot does not use a neural network to produce the graph representations. Instead, StreamSpot
It is a shallow method that uses an efficient hashing algorithm to encode each graph and defines anomaly score based on the similarity to graph clusters.
% based on Shingle-Vector frequency within the graph. The training phase of StreamSpot requires a bootstrap cluster file provided before the training starts to get initial cluster centroids of the normal graphs.

\textbf{DeepTraLog}
% The DeepTraLog model 
is a recently proposed deep heterogeneous graph-level anomaly detection method, which leverages request logs and heterogeneous graph data to learn an anomaly detection model. The log data is first modelled with natural language processing (NLP) model to acquire log embeddings and then combined with attributed trace graph data. 
% To ensure the training phase can be carried out in the same settings, 
Since our problem setting does not involve additional log embeddings, 
% we have simplified 
the DeepTraLog model using the graph-structured data only is taken in our experiment.
% and skipped the log representation learning part from its model architecture. Besides the log representation learning component, the graph representation is learnt using a GGNN-based deep SVDD model. The same GGNN-based model architecture is also implemented in our following experiments. Note that the detailed deep SVDD is implemented according to Section \ref{section:svdd}.

\textbf{GLocalKD}
% Global and Local Knowledge Distillation (GLocalKD) \cite{ma2022deep}
is a recent graph-level anomaly detection model on homogeneous graphs. It is adapted to heterogeneous graph data by replacing its homogeneous GCN architecture with the HetGCN architecture \cite{zhang2019heterogeneous}, with all other components unchanged.
% proposed to identify abnormalities in the structure and feature attributes of the graphs. It has been proven to be effective and robust around anomaly contaminations. Nonetheless, the original implementation of GLocalKD is developed to use homogeneous graphs, therefore, we have modified the current GLocalKD model accordingly to adopt the heterogeneous graph settings with different node and neighbourhood types. The minimal changes were made to the original GLocalKD's source code, and the critical difference is that the original \textit{GraphConv} layer is replaced with \textit{HetGraphConv} layer, which considers the neighbourhood types separately and aggregates all the neighbourhood types for node representation.

\textbf{OCHetGCN} is the baseline model we implement, in which the SVDD objective \cite{ruff2018deep} is applied to the HetGCN model to learn a one-class HetGCN model for heterogeneous graph-level anomaly detection.
% is implemented as described in Section \ref{section:baseline_hetgcn}.

The implementation details of our method and the four competing methods are presented in Appendix \ref{sup:implementation_details} \jxc{which could be used for reproducing the experiment results}.

% \subsection{Evaluation}
The model performance evaluation is carried out in two main categories, including the effectiveness of the model and the computational time cost. In terms of effectiveness, following the literature \cite{manzoor2016fast,pang2021toward,hezhe2023truncated}, two popular and complementary metrics, Area Under ROC Curve(AUC) and Average Precision (AP, also known as the area under the precision-recall curve), are used. Both metrics have a range of $[0,1]$; a larger AUC (AP) indicates better performance.
% in unsupervised anomaly detection experiments.
% when producing the model.

% \jxc{Moved Implementation Details into Appendix}

\subsection{Effectiveness on Real-world Datasets}\label{subsec:results}
 The model performance comparison details are described in Table \ref{tab:experiment_results}, and the best-performed results are highlighted in bold. According to this table, most of the baseline models performed quite similarly, where StreamSpot performed well in the FlowGraph dataset and the Simplified DeepTraLog performed well in the TraceLog dataset among all the baseline models. Furthermore, our proposed model HRGCN outperforms all the baselines in both AUC and AP and on both datasets. The average relative improvements in \% over the baseline models are around 20\% for the TraceLog dataset and 16.5\% for the FlowGraph dataset, respectively. It demonstrates that the proposed HRGCN framework is practical and produces better graph embeddings than the baselines.

\begin{table}
  \caption{Model Evaluation and Comparison Results.}
  \label{tab:experiment_results}
  \scalebox{0.9}{
  \begin{tabular}{c|cc|cc}
    \hline
    \multirow{2}{*}{\textbf{Model}} & \multicolumn{2}{c|}{\textbf{AUC} }  & \multicolumn{2}{c}{\textbf{AP}}  \\\cline{2-5}
          & \textit{TraceLog} & \textit{FlowGraph}  & \textit{TraceLog}  & \textit{FlowGraph} \\
    % \midrule
    \hline
    \texttt{OCHetGCN}& 0.628 &	0.952 &	0.518 &	0.954\\
    \texttt{GLocalKD}& 0.467 &	0.618 & 0.594 &	0.612\\
    \texttt{StreamSpot}& 0.634 &	\textbf{1.0} & 0.517	 &	0.992 \\
    \texttt{DeepTraLog} & 0.657 &	0.928 &	0.498 &	0.821 \\
    % \texttt{HRGCN}& 0.855 &	\textbf{1.0} & \textbf{0.762}	 &	\textbf{1.0}\\
    \texttt{HRGCN} (Ours)& \textbf{0.864} &	\textbf{1.0} & \textbf{0.754} &	\textbf{1.0}\\\hline
  \end{tabular}
  }
\end{table}

\subsection{Computational Efficiency}
% \jx{Done: Briefly address the limitation and ways to improve}
% \jx{Done: Give more clarity on why HRGCN have better inference time on larger graphs of the FlowGraph dataset where other methods typically run slower.}

Table \ref{tab:experiment_results_time} describes the training and predicting time cost in each model setup on both datasets. HetGCN took the least time to train a batch of instances on both TraceLog (batch of 8) and FlowGraph datasets (batch of 25). During predicting phase, StreamSpot performs the fastest in running a single instance on the TraceLog Dataset, while the Simplified DeepTraLog performs the fastest on the FlowGraph dataset. On the other hand, our proposed HRGCN model performs the slowest in both experiments. The HRGCN with Self-Supervised Module took 8.61 seconds and 54.71 seconds in training on TraceLog and FlowGraph datasets, respectively. However, during the predicting phase, HRGCN took only 1 second on TraceLog and 0.4 seconds on FlowGraph to run one instance. In addition, the training time for StreamSpot is skipped in this comparison because it uses a predefined list of bootstrap clusters and takes 0.072s and 0.006s to compute the initial cluster center for TraceLog and FlowGraph datasets. 
\jxc{It should be noted that HRGCN demonstrates faster inference on larger graphs within the FlowGraph dataset compared to smaller graphs in the TraceLog dataset. Despite the significantly larger graph sizes in the FlowGraph dataset, it only consists of 16 source-destination node type pairs (2 source node types and 8 destination node types), whereas the TraceLog graphs contain 64 pairs (8 for both source and destination node types). This highlights the dependence of computational efficiency on the number of source-destination node type pairs. \jxc{This situation could lead to a protracted training/inference phase when dealing with an extensive array of source-destination pairs within large-scaled graphs, unless effectively managed}. One potential improvement involves increasing parallelism by processing different relation type pairs simultaneously before the final aggregation step. 
% Nonetheless, this study will not focus on the details of the parallelism computations.
}
% \jxc{A detailed analysis of the computational time results is presented in Appendix \ref{sup:computational_efficiency_discussion}.}

\begin{table}
  \caption{Model Training and Predicting Time Results. Training and predicting time are the time spent running a single batch.}
  \label{tab:experiment_results_time}
  \scalebox{0.9}{
  \begin{tabular}{c|cc|cc}
    \hline
    \multirow{2}{*}{\textbf{Model}} & \multicolumn{2}{c|}{\textbf{Training Time} }  & \multicolumn{2}{c}{\textbf{Inference Time}}  \\\cline{2-5}
          & \textit{TraceLog} & \textit{FlowGraph}  & \textit{TraceLog}  & \textit{FlowGraph} \\
    % \midrule
    \hline
                    % & \textit{(8)} & \textit{(25)} & \textit{(1)} & \textit{(1)} \\
    % \midrule
    \texttt{OCHetGCN}&  \textbf{0.78s}  & \textbf{0.69s} & 0.054s  &  0.188s\\
    \texttt{GLocalKD}& 0.182s & 9.95s & 0.054s & 0.477s \\
    \texttt{StreamSpot}&  NA  &  NA  & \textbf{0.007s} & 6.212s \\
    \texttt{DeepTraLog} & 0.80s   & 0.727s & 0.032s &  \textbf{0.012s} \\
    % \texttt{HRGCN}&  4.62s & 9.01s & 1.048s &  0.381s \\
    \texttt{HRGCN} (Ours)& 8.61s  & 54.71s  & 1.057s & 0.395s \\\hline
  \end{tabular}
  }
\end{table}

\subsection{Ablation Study}
The proposed HRGCN introduces additional graph hierarchies into the framework and improves
graph representation learning. To understand the effectiveness of improving the model capability, we conduct ablation studies to evaluate the model performances of several variant settings, including a) baseline HetGCN model with introduced Heterogeneous Graph Data Augmentation (HetGDA); b) HRGCN model with only edge relation hierarchy, which removes source-destination relation hierarchy and HetGDA (HRGCN-ER); c) HRGCN model with only source-destination relation hierarchy, which removes edge relation hierarchy and HetGDA (HRGCN-SDR); d) HRGCN model with both edge and source-destination hierarchies but removes HetGDA (HRGCN-R2). Table \ref{tab:ablation} describes the performance results of each variant model. The results of the plain HetGCN model and the proposed HRGCN model are also logged to the table to make it easier to compare.

Table \ref{tab:ablation} demonstrates that the proposed HRGCN performs better than the models without considering the source-destination node type hierarchy. On the FlowGraph dataset, the HRGCN model with source-destination node type hierarchy (HRGCN-SDR) achieved optimal performance in both AUC and AP. On the TraceLog dataset, performance has improved significantly from the HRGCN-SDR model variant,  which achieved around 25\% boost in both AUC and AP (compared to the plain HetGCN model). HRGCN model with edge-type relation hierarchy (HRGCN-ER) has a limited effect over the FlowGraph dataset, which increases AUC by 3\% while no improvement in AP. On the other hand, the HRGCN-ER slightly improves the performance on the TraceLog dataset by 6\% and 2\% in AUC and AP, respectively. Although both individual relation hierarchies made some improvements compared to the plain HetGCN, combining both relation hierarchies (HRGCN-R2) did not receive a significant performance boost. This is also reflected in the full HRGCN model performance. Nonetheless, these three model variants demonstrate performance trade-offs between AUC and AP, i.e., \textit{HRGCN-SDR} performs the best in AUC but has lower AP; \textit{HRGCN-R2} performs the best in AP but has lower AUC; \textit{HRGCN(full)} shows a balanced performance of AUC and AP. In addition, applying Heterogeneous Graph Data Augmentation alone with the plain HetGCN model did not help the TraceLog data. Still, it significantly improved the model performance by 5\% when applied to the FlowGraph dataset.

\begin{table}
    \caption{Model Ablation Analysis on Multiple Variants.}
    \centering
    \scalebox{0.9}{
    \begin{tabular}{c|cc|cc}
    \hline
    \multirow{2}{*}{\textbf{Model}} & \multicolumn{2}{c|}{\textbf{AUC} }  & \multicolumn{2}{c}{\textbf{AP}}  \\\cline{2-5}
          & \textit{TraceLog} & \textit{FlowGraph}  & \textit{TraceLog}  & \textit{FlowGraph} \\
    % \midrule
    \hline
    \texttt{OCHetGCN(base)} & 0.628 &	0.952 &	0.518 &	0.954 \\
    \texttt{OCHetGCN-HetGDA} & 0.573 &	\textbf{1.0} &	0.452 &	\textbf{1.0} \\
    \texttt{HRGCN-ER} & 0.683 &	0.988 &	0.534 &	0.953 \\
    \texttt{HRGCN-SDR} & \textbf{0.871} &	\textbf{1.0} &	0.743 &	\textbf{1.0} \\
    \texttt{HRGCN-R2} & 0.858 &	\textbf{1.0} &	\textbf{0.761} &	\textbf{1.0} \\
    \texttt{HRGCN (full)} & 0.864 &	\textbf{1.0} &	0.754 &	\textbf{1.0} \\\hline
    \end{tabular}
    \label{tab:ablation}
    }
\end{table}

\begin{figure}
    \centering
    \scalebox{0.9}{
    \includegraphics[width=\linewidth]{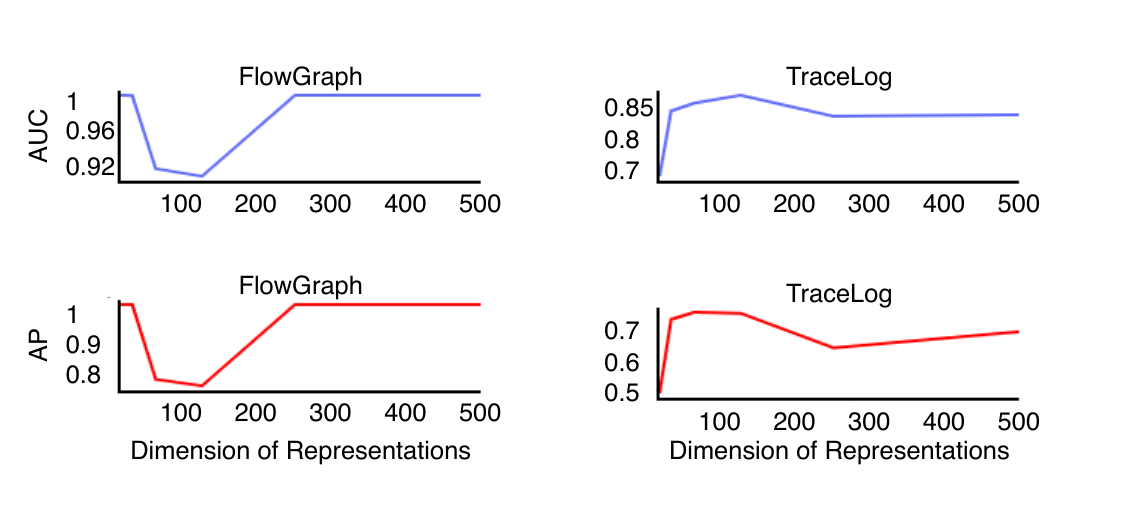}
    }
    \caption{Results of Varying Representation Dimensions.}
    \label{fig:hyper_analysis_channels}
\end{figure}

\begin{figure}[ht!]
    \centering
    \scalebox{0.9}{
    \includegraphics[width=\linewidth]{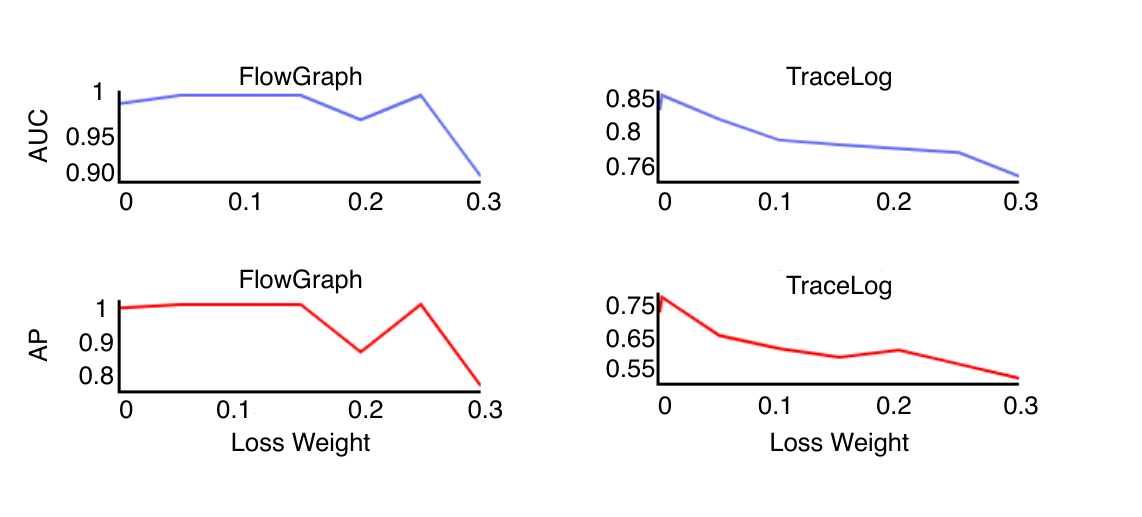}
    }
    \caption{Impact of Self-Supervised Loss in HRGCN Model.}
    \label{fig:hyper_analysis_loss}
\end{figure}

\subsection{Hyperparameter Analysis}
Hyperparameters play an essential role when training HRGCN on different datasets, such as the size of the hidden layer and the weights of self-supervised loss.
% Several experiments are conducted to analyse the impacts of the three hyperparameters.
%In addition, the same set of analyses is carried out on both datasets.

Figure \ref{fig:hyper_analysis_channels} illustrates the impact of the representation dimension sizes. Both evaluation metrics are more consistent when the dimension size increases or is set to a relatively smaller number. From the FlowGraph dataset, the model returns a more consistent performance when the hidden layer size is larger than 256. In comparison, the model performs the best when the dimension size is set to 128 on the  TraceLog dataset.

The results of varying self-supervised loss weights is shown in Figure \ref{fig:hyper_analysis_loss}.
%and \ref{sup:hyperparameter_analysis_layers}, respectively. 
On the FlowGraph dataset, the model achieves the optimal performance when the loss weight is smaller than 0.15, while the TraceLog dataset requires a narrower range between 0 and 0.01.
% In addition, both datasets receive optimal model performance when the number of hidden layers is 2.

\section{Industrial Case Study}\label{case_study}
%With the industrial setting in the real world, we have also applied the HRGCN model to a close to real-world scenario.

\subsection{Problem Statement}
This section will introduce the details of the problem we try to tackle in real-world industrial settings, including the problem definition, hypothesis, and the dataset used for conducting this study.

\subsubsection{Problem Definition}
A typical network behaviour can be described in combination with entities of Users, Cells, and Sectors. Users are the objects that perform or receive certain network actions that happen to themselves. Cells are the most granular network devices (in this scenario) that serve certain network actions to individual users. Sectors describe a collection of Cell devices and contain the logical grouping information among every Cell device (see Figure \ref{fig:case_study_data_sup} (a) in Appendix \ref{sup:case_study_dataset_overview} for a simplified example of how the network graph would look).

Serving good connections and networks to customers is always one of the goals of a telecommunication company. Therefore, in this study, we are trying to detect if any anomalous behaviours are happening within the network that could impact user's experience. In this case, we can identify the specific network devices and times they happened and provide a means for monitoring and post-event analysis of the network behaviour.

Given a collection of the network event data $\mathcal{D} \sim \left\{n_0, n_1, ..., n_n\right\}$ that happened in a certain period $T$. We expect that there exist certain relations within the network events which could be modelled by a differentiable function $\mathcal{S}$, which outputs an anomaly score $P$ for the input network events. Note that an anomaly score of 1 means highly anomalous, and 0 indicates no detected anomalous patterns:

\begin{equation}
    P(1|d,t) = \mathcal{S}(d, t), \{d \in \mathcal{D}; t \in T\}
\end{equation}

\subsubsection{Trace Event Dataset} \label{sec:case_study_dataset}

% \jx{TODO: Add brief details on the security and privacy consideration of the dataset, including how/why this is being sampled.}

The Trace Event Dataset is populated based on relevant mobile event network data at TPG Telecom in Australia. We have simulated and resampled the new Trace Event Dataset with similar data statistics and properties so that this could reflect closely to real-world events. 

To keep an appropriate data size to conduct a case study, we have sampled 20\% of simulated network event data in a specific period. The sampled network data contains in total of 79,446,610 raw network events. Each network event can be categorised into one of 30 individual network event codes, and the frequency of each event that happened throughout the day could vary significantly. For example, the most frequent event could occur 49 million times, while the least frequent event may occur only two times. Moreover, some graphs may contain many vertices in a certain node and neighbourhood type (see Figure \ref{fig:case_study_data_sup} (b) in Appendix \ref{sup:case_study_dataset_overview} for an example of how the relation between User and Cell would look like).

After processing the raw data into graph structures, we formed the dataset containing 21,849 traces in formatted graph data around the network events that happened to users on each network device to track the Sector behaviours. There are 876 abnormal graphs in the dataset, with around 4\% occurrence in this dataset. Besides, the average-sized traces consist of around 478 nodes and 477 edges individually. The dataset has three node types: User, Cell, and Sector. There are around 2.5 million user nodes, 26 thousand distinct Cell nodes, and 9 thousand distinct Sector nodes. Because the dataset is formed based on different periods, the total number of nodes is around 10 million. A detailed summary of the data statistics can be found in Table \ref{tab:case_study_data} in Appendix \ref{sup:case_study_dataset_overview}.
% summarises the details of the data properties and their basic statistics. 

Furthermore, although we are conducting the unsupervised training methodology, we still need predefined labels to evaluate the model performances. In this case, we also introduce SME (Subject Matter Expert) knowledge onto the Trace Event Dataset to help with the evaluation purposes.

The following experiment divides the dataset into training, validation, and testing sets. The training and validation set contains 60\% and 20\% of normal traces, and the rest normal and abnormal traces are grouped into the testing set.

%12,619 normal traces, the evaluation set contains 4,611 traces (4,175 normal and 436 abnormal), and the testing set contains 4,619 traces (4179 normal and 440 abnormal).

\subsection{Experiment \& Results}
% In this section, the details of the experiments are provided, such as the hyperparameter used for the final model, training steps, and evaluation results. 
% \jxc{Hyperparameters and hardware settings are described in Supplement \ref{sup:case_study_implementation}.}

% \subsubsection{Results}\label{sec:industrial_result}

Figure  \ref{fig:case_study_result_test} illustrates the model performance comparison on the test dataset. The proposed HRGCN model outperforms the other two models by 5\% in AUC and 37\% in AP. 
% (see Supplement \ref{sup:industrial_evaluation} for the performance changes over epochs). 
We found that the model could converge at an early stage and performed stably for the rest of the training iterations.

\begin{figure}
    \centering
    \scalebox{0.9}{
    \includegraphics[width=\linewidth]{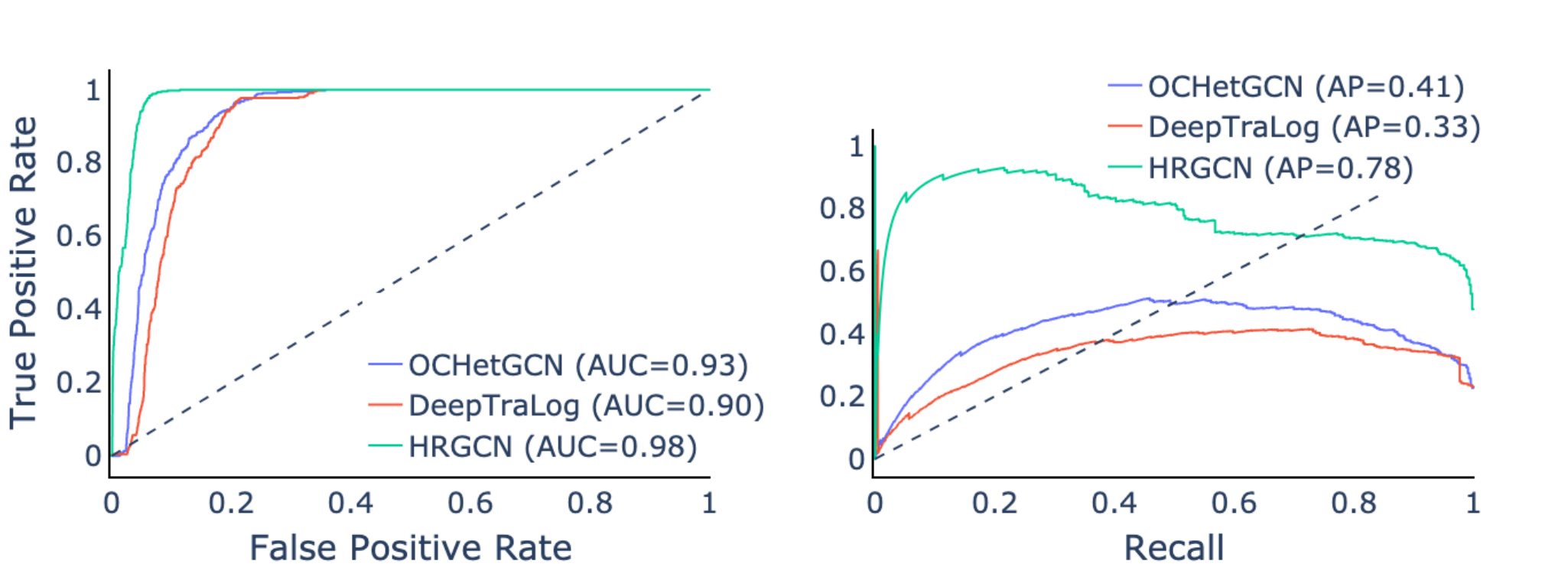}
    }
    \caption{Testing Results on the Industrial Trace Event Data}
    \label{fig:case_study_result_test}
\end{figure}

% \subsection{Analysis \& Discussion}
To understand the effectiveness of the produced model, we have also applied the model to an unseen set of traces for the network event. In this section, we'll dive deep into the model outputs and the reflections from the results.

\subsubsection{Trace Representation \& Anomaly Scores}
Since the model's output produces vectors in 300 dimensions, we applied additional post-processing to the model outputs to visualise the trace representation onto a 2D plane which projects each trace representation into lower dimensions. 
% For example, we combined PCA and T-distributed Stochastic Neighbor Embedding (t-SNE) \cite{ljpvd2008visualizing} to reduce the vector dimension to 2. 
Figure \ref{fig:case_study_scores_viz} illustrates the scatter view of each trace in test sets, and it shows that most abnormal trace representations are closely grouped into a simple space throughout the 2D plane. 
% On the other hand, Figure \ref{fig:case_study_scores_hist} in Supplement \ref{sup:industrial_evaluation_hist} describes the distribution of the predicted anomaly scores 
% across the test traces
. most normal traces have a score very close to 0.0025, while most abnormal traces have a score of 0.15. %In addition, drawing a certain threshold, such as a score = 0.06, clearly distinguishes between most normal and abnormal traces in the testing dataset. Furthermore, 

\begin{figure}
    \centering
    \scalebox{0.9}{
    \includegraphics[width=\linewidth]{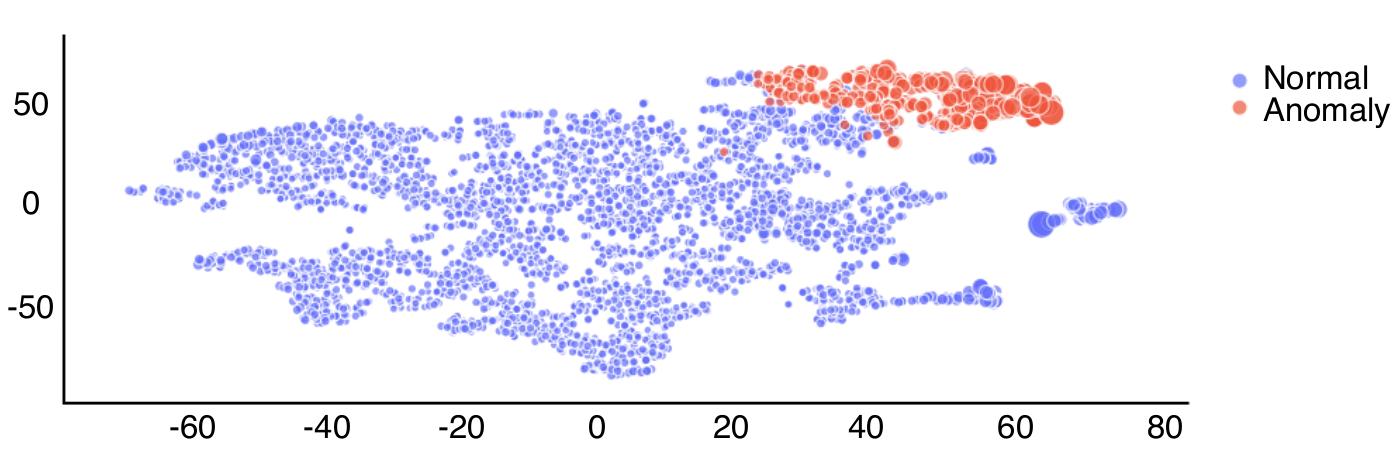}
    }
    \caption{t-SNE visualisation of graph representation on the unseen trace data. }
    % Each data point is a trace representation.}
    \label{fig:case_study_scores_viz}
\end{figure}

\subsubsection{Network Analysis}

\jxc{Figure \ref{fig:case_study_data} (a) and (b) illustrate a simplified network structure of how a normal/abnormal trace would look like. Although the trace graph structure may be similar, individual node features may be evolved over time and certain devices may become anomalous.}

\begin{figure}
    \centering
    \scalebox{0.8}{
        \includegraphics[width=\linewidth]{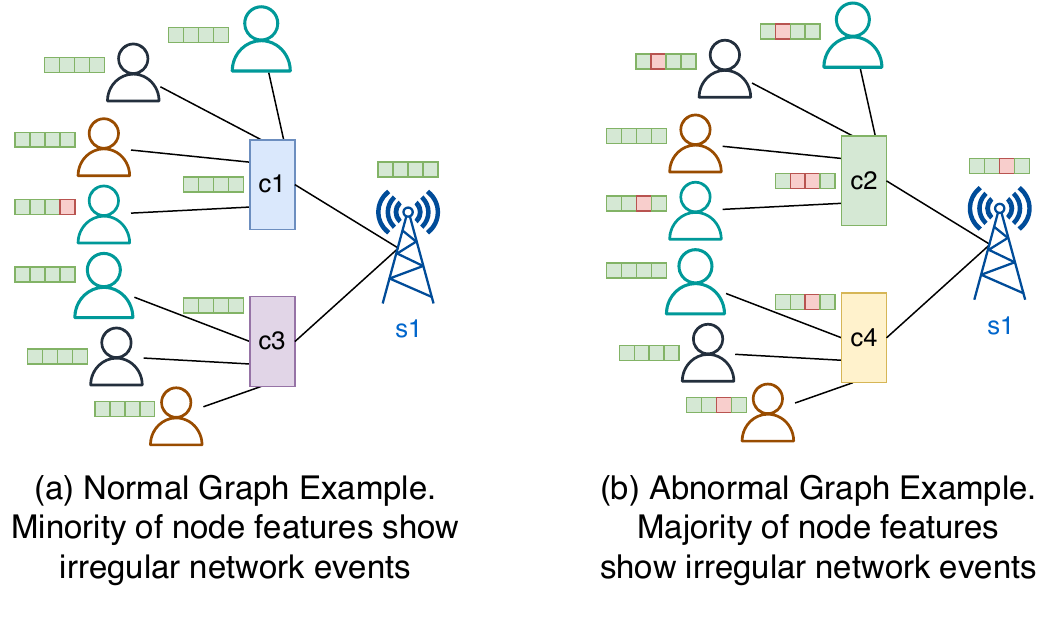}
    }
    \caption{Example Anomalies in Trace Event Graph. (a) and (b) demonstrate simplified examples of normal/abnormal graphs. Individual node features may reflect on the overall performance of the network device. In simple words, high volume of users received irregular network events demonstrates there is a congested network device, low volume of users received irregular events may not indicate a malfunctioning device.}
    \label{fig:case_study_data}
\end{figure}

To investigate deeper onto the network analysis around the device performances and experiences, we have sampled a collection of devices and started to look closer at their results for monitoring and alerting purposes. To do so, we have rearranged the view to demonstrate the anomaly score changes within certain sequential time windows. Figure \ref{fig:case_study_devices_abnormal} demonstrates that for the selected abnormal devices that experienced congested behaviours. 
% In comparison, Figure \ref{fig:case_study_devices_normal} in Supplement \ref{sup:normal_device_over_time} demonstrates a relatively flat score changes over time on sampled normal devices. 
From the results, we can see that the model is adaptive to behaviour changes on individual devices over time and can detect and identify intermittent normal/abnormal behavioural changes throughout network devices.

\begin{figure}
    \centering
    \scalebox{0.9}{
    \includegraphics[width=\linewidth]{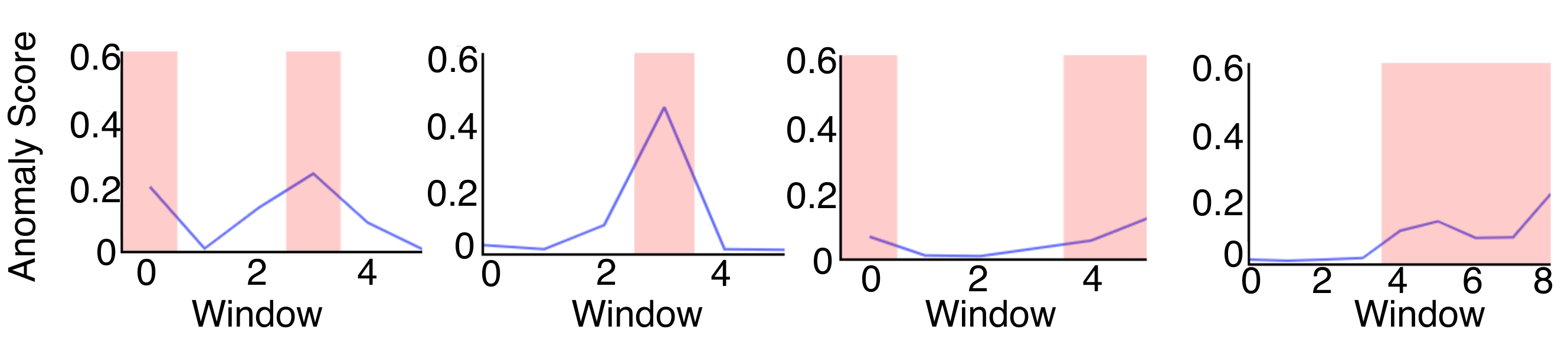}
    }
    \caption{Anomaly Scores Over Time on Abnormal Devices }
    \label{fig:case_study_devices_abnormal}
\end{figure}

% \jx{Done. Added elaborations on the details of the anomaly times, related to sample graphs to illustrate (In Figure 4}

\section{Conclusion}\label{conclusion}

In this work, we propose an unsupervised model, namely Hierarchical Relation-augmented Heterogeneous Graph Convolutional Neural Network (HRGCN) to improve the representation learning task on heterogeneous graph (HG) data. We improve the effectiveness of modelling additional relations between edges and source-destination node neighbourhoods to discover deep latent information within the complex heterogeneous graph networks without labels. Furthermore, a complementary self-supervised learning module and associated Heterogeneous Graph Data Augmentation (HetGDA) method are also introduced to enhance the model's performance during training. Extensive experiments on multiple system-focused anomaly detection datasets demonstrate that the HRGCN model can significantly outperform state-of-the-art methods. In addition, the impact of each proposed module, i.e., the relation-hierarchical module and self-supervised module, is unravelled to demonstrate the performance boosts.

On the other hand, an industrial case study is conducted over a real-world industrial setting in telecommunication. 
% The case study includes the details of the dataset description, model training aspects, model performance analysis, and device congestion analysis. In the end, 
We demonstrate through the case study that our proposed HRGCN model can work effectively in real-world settings and significantly outperforms current state-of-the-art models.
% and its effectiveness against real-world prediction tasks.

% For future work, \textcolor{blue}{talk about improvement to training times and efficiency. Additional research into the augmentation of the hierarchical relations in HG}

% Removed for double-blind reviews
\section*{Acknowledgment}

This research is supported by TPG Telecom Australia. Any conclusions drawn in this study are of the authors and do not necessarily reflect the views and perspectives of TPG Telecom. 
% \gs{it is double-blind review; need to remove this section.}

\bibliography{references}
\bibliographystyle{IEEEtran}

\newpage

\phantom{placeholder}

\newpage

% \vspace{12pt}

\appendix
% \section{Supplementary Materials}

\subsection{TraceLog and FlowGraph Dataset Overview}\label{sup:dataset_overview}
The detailed statistics of the TraceLog and FlowGraph datasets are summarised in Table \ref{tab:dataset_overview}.

\begin{table}[htbp]
    \caption{TraceLog and FlowGraph Dataset Overview}
    \centering
    \begin{tabular}{l|l}
    \hline
         \textbf{TraceLog} & \textbf{FlowGraph} \\
    \hline
       \texttt{\#node-types:} 8  & \texttt{\#node-types:} 8 \\
       \texttt{\#edge-types:} 4  & \texttt{\#edge-types:} 26 \\
       \texttt{\#node-attributes:} 7  & \texttt{\#node-attributes:} 26 \\
       \texttt{\#graphs:} 132,485  & \texttt{\#graph:} 600 \\
       \texttt{\#anomaly-graph:} 23,334  & \texttt{\#anomaly-graph:} 100\\
       \texttt{\#mean-size:} 205  & \texttt{\#mean-size:} 8,411 \\
       \texttt{\#mean-edges:} 224  & \texttt{\#mean-edges:} 12,730 \\
       \texttt{\#mean-node-types:} 8  & \texttt{\#mean-node-types:} 8 \\
       \texttt{\#mean-edge-types:} 2.87  & \texttt{\#mean-edge-types:} 25.5  \\
   \hline
    \end{tabular}
    \label{tab:dataset_overview}
\end{table}

\subsubsection{Details of TraceLog Dataset}\label{sup:tracelog_details}
Specifically, the dataset has 132,485 trace graphs, including 23,334 anomalous graphs and 109,151 normal graphs. 
% It is split into training and testing datasets, where training data contains 
% Among the normal graphs, around 
60\% and 20\% normal graphs are used as training and validation datasets. The remaining normal graphs are combined with abnormal graphs as the test data. The dataset contains 8 node types and 4 edge types in total. On average, each graph consists of around 205 nodes, 224 edges, 8 node types and 2.87 edge types. 

\subsubsection{Details of FlowGraph Dataset}\label{sup:flowgraph_details}
There are 8 node types and 26 edge types in total. On average, each graph contains around 8,411 nodes, 12,730 edges, 8 node types, and 25.5 edge types. 
% The details could be also found in Table \ref{tab:dataset_overview}. \gs{what is the attack?}\jx{Added above}
% \gs{how many normal and abnormal graphs we have in this dataset in total?}\jx{Added in above}. 
FlowGraph is also split into training, validation, and testing. Following \cite{manzoor2016fast}, 
% inherited a similar training graph setting from StreamSpot \cite{manzoor2016fast}, which uses around
60\% and 15\% normal graphs are used for training and validation, and the rest is combined for the testing set.

\subsection{Implementation Details}\label{sup:implementation_details}
% \textcolor{blue}{[Describe the hyperparameter used for both datasets]}

% \gs{add the details of GNN architectures, optimisation hyperparameters, etc. for all models}. \jx{Details added, also put them all in Table \ref{tab:hyperparameters_FlowGraph} and \ref{tab:hyperparameters_tracelog} in Supplement \ref{sup:hyperparameters}}

The following hyperparameters are used 
% to achieve the optimal performance of
in the proposed HRGCN. On the TraceLog dataset, the learning rate is 0.0001, the representation dimension is 300, the number of hidden layers is 2, the self-supervised loss weight is 0.001, and the batch size is 8. In addition, the parameters used for heterogeneous graph data augmentations are: edge perturbation probability is 0.84, edge replacement percentage is 0.13, swap node type percentage is 0.1,  and swap edge type percentage is 0.17. On the FlowGraph dataset, the learning rate is 0.01, the representation dimension is 32, the number of hidden layers is 2, the self-supervised loss weight is 0.21, the batch size is 25, the edge replacement percentage is 0.39, and swap node type percentage is 0.52. 
During the training process, the model is trained on the training set, and tuned on the validation set using the self-supervised classification accuracy and SVDD distance. 
A network architecture of the same depth and breadth is used in all deep competing methods. The implementation of StreamSpot is taken from its authors.
% More details of the hyperparameters used in the competing models are described in Supplement \ref{sup:hyperparameters}.

% \textbf{Hyperparameters for OCHetGCN}. On the TraceLog dataset, the learning rate is set to 0.0001, the representation dimension is set to 300, the number of hidden layers is set to 2, and the batch size is set to 128. While a learning rate of 0.01, a representation dimension of 32, the number of hidden layers is set to 2, and a batch size of 25 are set for the FlowGraph dataset.

% \textbf{Hyperparameters for GLocalKD}. On the TraceLog dataset, the learning rate is set to 0.0001, the representation dimension is set to 300, and the batch size is set to 128. While a learning rate of 0.01, a representation dimension of 32, and a batch size of 4 are set for the FlowGraph dataset.

% \textbf{Hyperparameters for StreamSpot}. On both datasets, the chunk length is set to 60, sketch size is set to 1000.

% \textbf{Hyperparameters for DeepTraLog}. On the TraceLog dataset, the learning rate is set to 0.0001, the representation dimension is set to 300, and the batch size is set to 128. While a learning rate of 0.01, a representation dimension of 32, and a batch size of 4 are set for the FlowGraph dataset.

% \textbf{Hyperparameters for HRGCN}. On the TraceLog dataset, the learning rate is set to 0.0001, the representation dimension is set to 300, and the batch size is set to 8. While a learning rate of 0.01, a representation dimension of 32, and a batch size of 25 are set for the FlowGraph dataset.

% \subsubsection{Hardware}
The experiment is conducted in AWS using SageMaker Training Jobs\footnote{https://aws.amazon.com/sagemaker/}. The instance type used for training is g4dn.2xlarge, which has 8 vCPU, 32 GiB memory, and 1 NVIDIA T4 GPU. 

\subsection{Case Study -- Overview of the Trace Event Dataset}\label{sup:case_study_dataset_overview}
The properties and statistics of the case study dataset is summarised in Table \ref{tab:case_study_data}.

An example of trace event graph example and schema is presented in Figure \ref{fig:case_study_data_sup}.

\begin{table}[htbp]
    \caption{Case Study Dataset Properties and Statistics}
    \centering
    \begin{tabular}{l|l|l}
    \hline
        \textbf{Trace} & \textbf{Node} & \textbf{Edge}  \\
        \hline
        \texttt{\#traces:} 21,849 &\texttt{\#node-types: 3} & \texttt{\#edge-types:} 30 \\
        \texttt{\#normal:} 20,973 & \texttt{\#total:} 10,423,149 & \texttt{\#total:} 10,423,149 \\
        \texttt{\#abnormal:} 876 & \texttt{\#user:} 2,597,531  &   \\ 
        \texttt{mean-size:} 478 & \texttt{\#cell:} 26,761  &   \\ 
        \texttt{\#mean-edges:} 477& \texttt{\#sector:} 8,724  &  \\
    \hline
    \end{tabular}
    \label{tab:case_study_data}
\end{table}

\begin{figure}
    \centering
    \includegraphics[width=\linewidth]{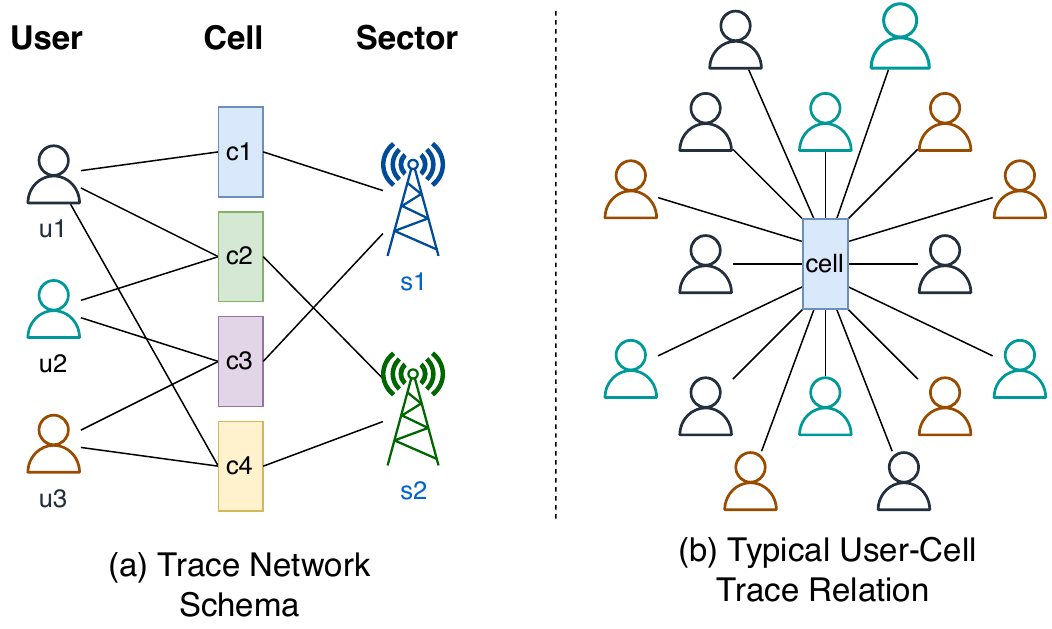}
    \caption{Trace Event Graph Example and Schema. (a) demonstrates the basic relationships among User, Cell, and Sector. (b) illustrates how a typical Cell would look like in the abstract when there is a lot of network traffic.}
    \label{fig:case_study_data_sup}
\end{figure}

\subsection{Case Study Implementation details}\label{sup:case_study_implementation}
% \subsubsection{Hyperparameters}
We have selected a set of hyperparameters for the Trace Event Datasets based on the previous experiments on the FlowGraph and TraceLog datasets. The learning rate is set to 0.01, the hidden layer size and output embedding size are 300,  the number of hidden HetGCN layers is  2, and the batch size is 256. The experiments are conducted in the same environment as that on the other two datasets.

\end{document}